\documentclass[10pt,twocolumn,letterpaper]{article}
\usepackage[accsupp]{axessibility}

\usepackage{wacv}
\usepackage{times}
\usepackage{epsfig}
\usepackage{graphicx}
\usepackage{amsmath}
\usepackage{amssymb}
\usepackage{booktabs}
\usepackage{multirow}

\usepackage{amssymb}
\usepackage{pifont}
\newcommand{\cmark}{\ding{51}}%
\newcommand{\xmark}{\ding{55}}%
\usepackage{color, colortbl}

\definecolor{Gray}{gray}{0.9}

\def\wacvPaperID{335} 

\wacvfinalcopy 

\ifwacvfinal
\usepackage[breaklinks=true,bookmarks=false]{hyperref}
\else
\usepackage[pagebackref=true,breaklinks=true,colorlinks,bookmarks=false]{hyperref}
\fi

\begin{document}

\title{STAR-Transformer: A Spatio-temporal Cross Attention Transformer for Human Action Recognition}

\setcounter{footnote}{0}
\author{Dasom Ahn\textsuperscript{1}, Sangwon Kim\textsuperscript{1}, Hyunsu Hong\textsuperscript{2}, Byoung Chul Ko\textsuperscript{1}\thanks{Corresponding author}\and
\textsuperscript{1}Dept. of Computer Engineering, Keimyung University, Daegu, South Korea\\
\textsuperscript{2}Difine, Seongnam, South Korea\\
{\tt\small \{ektha772, eddiesangwonkim\}@gmail.com, phil@difine.co.kr, niceko@kmu.ac.kr}
}

\maketitle
\thispagestyle{empty}

\begin{abstract}
    In action recognition, although the combination of spatio-temporal videos and skeleton features can improve the recognition performance, a separate model and balancing feature representation for cross-modal data are required. To solve these problems, we propose Spatio-TemporAl cRoss (STAR)-transformer, which can effectively represent two cross-modal features as a recognizable vector. First, from the input video and skeleton sequence, video frames are output as global grid tokens and skeletons are output as joint map tokens, respectively. These tokens are then aggregated into multi-class tokens and input into STAR-transformer. The STAR-transformer encoder consists of a full spatio-temporal attention (FAttn) module and a proposed zigzag spatio-temporal attention (ZAttn) module. Similarly, the continuous decoder consists of a FAttn module and a proposed binary spatio-temporal attention (BAttn) module. STAR-transformer learns an efficient multi-feature representation of the spatio-temporal features by properly arranging pairings of the FAttn, ZAttn, and BAttn modules. Experimental results on the Penn-Action, NTU-RGB+D 60, and 120 datasets show that the proposed method achieves a promising improvement in performance in comparison to previous state-of-the-art methods.
\end{abstract}

\section{Introduction}

Action recognition is a traditional research topic that classifies human actions using video frames and has been applied in various applications, including human-robot interaction \cite{bandi2021skeleton}, healthcare \cite{serpush2022wearable}, and video surveillance \cite{elharrouss2021combined}. With the recent development of deep learning, action recognition research trends have been divided into three approaches. First, in a video-based approach \cite{zhu2018action,baradel2018glimpse,wang2021action,liu2021no,guo2014survey,mazzia2022action,girish2020understanding}, deep learning models use only video frames to recognize the action. This approach results in a significant degradation in performance owing to various noises from the wild, such as differences in the camera angles and sizes of the human targets and complex backgrounds. The second is a skeleton-based approach \cite{thakkar2018part, caetano2019skelemotion, cheng2020skeleton, duan2022revisiting, tran2018closer, feichtenhofer2020x3d}. 
\begin{figure}[t]
\begin{center}
   \includegraphics[width=1.0\linewidth]{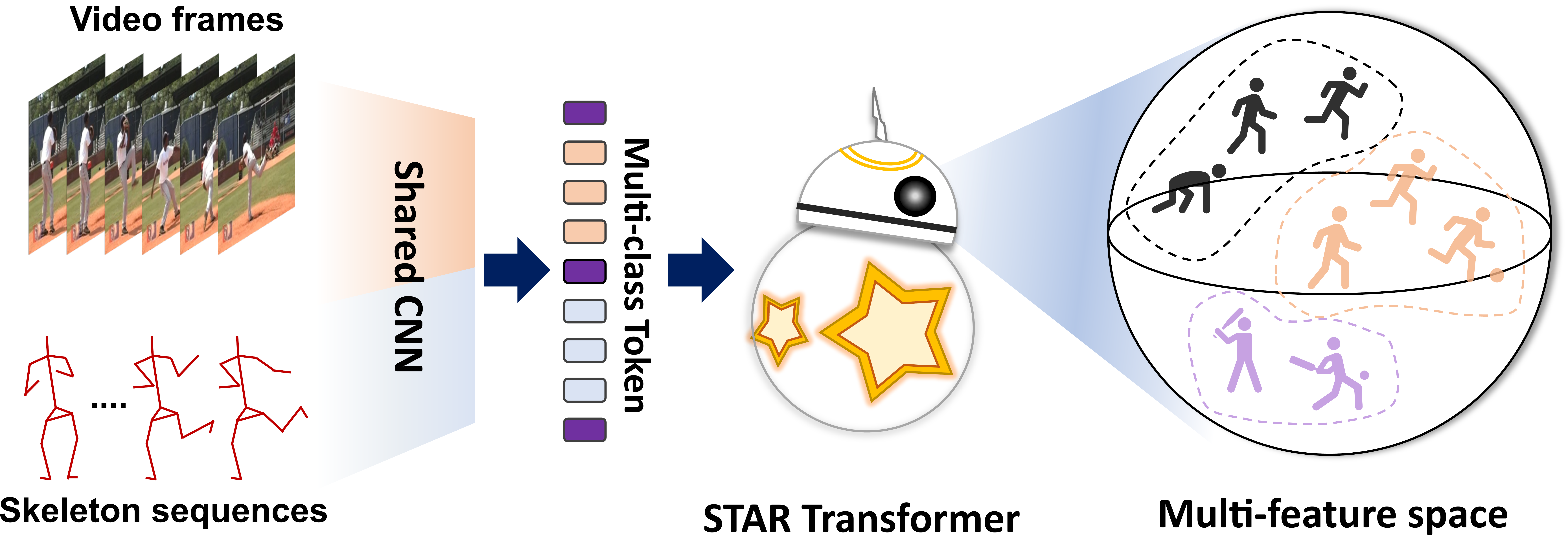}
\end{center}
   \caption{The proposed STAR-transformer takes a multi-class token as input and transforms it into a class-separable multi-feature representation. The multi-class token is an aggregation of global grid and joint map tokens obtained by feeding video and pose sequences into a shared CNN.}
   \vskip -0.5cm
\label{fig1}
\end{figure}
Actions are recognized using human skeletons and joint trajectories in different time zones as inputs into the deep learning model. However, an additional deep learning model is required to extract the human skeleton from an image. In addition, the action recognition is largely dependent on the accuracy of the skeleton extractor and the degree of overlap of the skeleton.
The third approach is the use of cross-modal data, video, and skeletons together \cite{davoodikakhki2020hierarchical, das2020vpn, su2020msaf}. A deep learning model learns the RGB of the video frames and human skeletal features together; thus, it generally shows a high recognition performance. However, combining video and skeleton data is an ambiguous process and requires a separate submodel for cross-modal learning.

As a new learning paradigm in the deep learning field, Vision Transformer (ViT) \cite{dosovitskiy2020image} has recently attracted attention owing to its excellent performance in various computer vision fields such as image classification \cite{swkim2022vitnet}, image segmentation \cite{yan2022after}, object tracking \cite{zeng2021motr}, and action recognition \cite{chen2022mm}. The self-attention mechanism, which is a key element of ViT, is specialized for determining the spatial relationship of each image and is effectively applied to image classification. However, in action recognition, the features of long-range frames and multiple features that change over time must both be considered; therefore, ViT based on the existing multi-head attention mechanism has limitations in terms of a high computational cost \cite{liang2021dualformer}.

In this study, we propose a multi-feature representation method based on cross-modal learning and a Spatio-TemporAl cRoss transformer (STAR-transformer) attention mechanism. For cross-modal learning, we propose a method of aggregating the cross-modal data of spatio-temporal video and a skeleton into a multi-class token to solve the problem of combining cross-modal action data. STAR-transformer consists of a new cross-attention module that replaces the multi-head attention of a simple ViT. The proposed STAR-transformer has demonstrated an excellent performance through various experiments.

Figure \ref{fig1} shows the overall operational structure of STAR-transformer. Two cross-modal features are fed into the shared convolutional neural network (CNN) model and separated into multi-class tokens. STAR-transformer consists of an $L$-layer encoder–decoder output separable multi-class feature, which is used as input for the downstream action recognition network.

The contributions of this paper can be summarized as follows.

$\bullet$ Cross-modal learning: It is possible to flexibly aggregate spatio-temporal skeleton features as well as video frames and effectively learn cross-modal data to create multi-class tokens.

$\bullet$ STAR-transformer: The existing self-attention mechanism is limited to the application of action recognition because it focuses on the relation between spatial features. We therefore propose a STAR attention mechanism that can learn cross-modal features. The encoder and decoder of STAR-transformer are composed of zigzag and binary skip STAR attention.

$\bullet$ Various performance evaluation experiments: A performance evaluation was conducted based on several benchmark datasets, and the proposed model showed a better performance than existing state-of-the-art (SoTA) models.

\section{Related Work}

{\bf Video and image-based action recognition :} It aims to recognize actions using only sequential \cite{zhu2018action, baradel2018glimpse, wang2021action, liu2021no, tran2018closer, feichtenhofer2020x3d} or still images \cite{guo2014survey, girish2020understanding}. The general process of video-based action recognition involves breaking the action into smaller semantic components and understanding the importance of each component in action recognition \cite{girish2020understanding}. Because this method uses video frames, it can be processed using a simple single model. However, if the video is long, the recognition speed will be slow and the performance will be significantly affected by the various noises from the wild.

{\bf Skeleton-based action recognition:} It aims to recognize actions by applying a list of spatio-temporal joint coordinates of video frames extracted from a pose estimator to a graph convolutional network (GCN) \cite{thakkar2018part, cheng2020skeleton, cai2021jolo}, 3D-CNN \cite{duan2022revisiting}, and CNN \cite{caetano2019skelemotion}. A skeleton sequence has an advantage of being unaffected by contextual disturbances such as changes in background and lighting \cite{duan2022revisiting} but has a disadvantage in that the recognition performance is largely dependent on the pose extractor and requires an extra classifier for recognition.

{\bf Video and skeleton-based action recognition:} It aims to achieve a high action recognition performance by fusing multi-modal (cross-modal) information into an integrated set of discriminative features \cite{davoodikakhki2020hierarchical, das2020vpn, su2020msaf, joze2020mmtm}. The video-pose network (VPN) action recognition mechanism \cite{das2020vpn}, which uses cross-modal features and knowledge distillation to infuse poses into RGB streams, has proven that cross-modal features can achieve a better performance than unimodal features. Despite its relatively high recognition performance, this method still has problems in the design of a subnetwork for cross-modal learning and methods for combining cross-modal data.

{\bf Transformer-based action recognition:} Because a transformer is a powerful tool in terms of long-range temporal modeling when using a self-attention module \cite{arnab2021vivit}, an increasing number of studies in this area, particularly action recognition, have been conducted \cite{arnab2021vivit, bertasius2021space, tong2022videomae}. Most action recognition approaches using a transformer apply video frames as input tokens \cite{wang2021oadtr, girdhar2019video, xu2021long, zhang2021co, yi2021asformer, mazzia2022action}, and relatively few methods use the skeleton \cite{shi2020decoupled, plizzari2021skeleton, mazzia2022action} of the transformer. However, transformer-based action recognition often suffers from high computational costs owing to self-attention given to the large number of 3D tokens in a video \cite{liang2021dualformer}. Moreover, an approach to coupling cross-modal information using a transformer has yet to be developed. Therefore, this study is the first attempt at using spatio-temporal cross-modal data as input tokens for ViTs without applying separate sub-models.

\begin{figure*}
\begin{center}
    \includegraphics[width=1.0\linewidth]{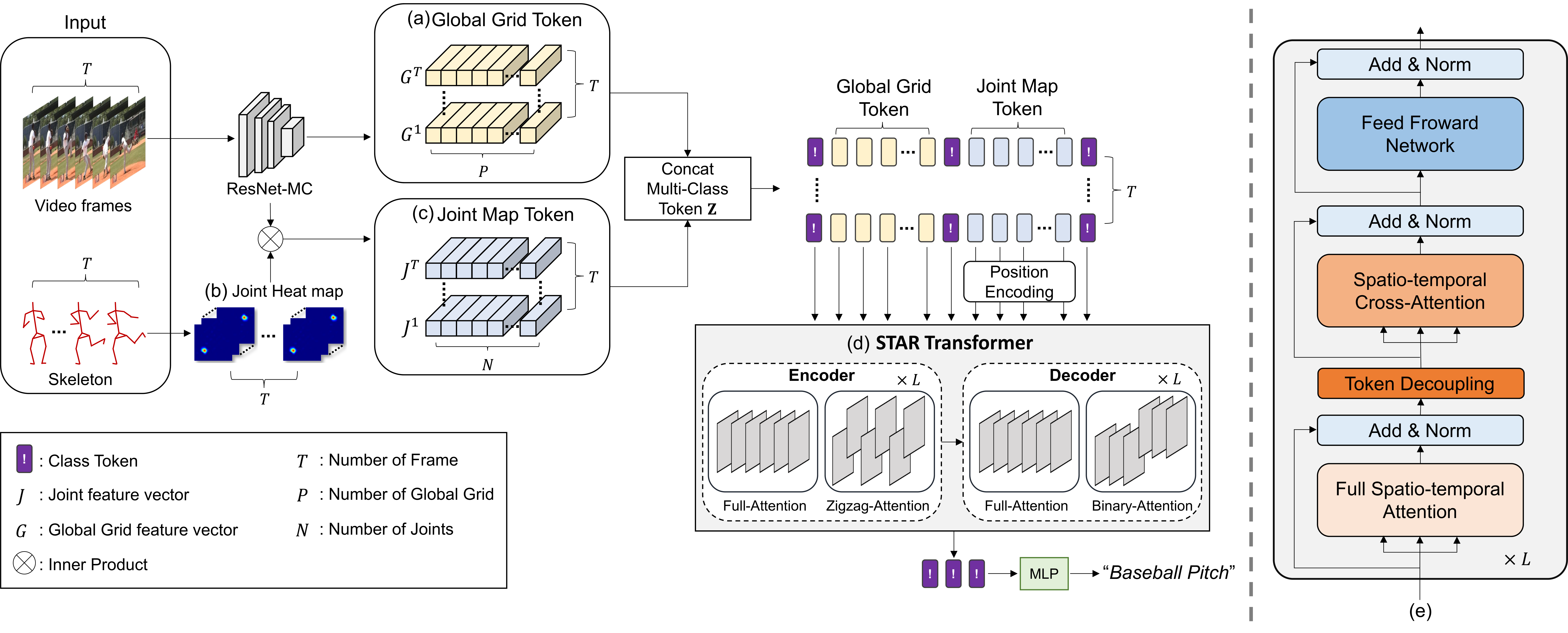}
\end{center}
   \caption{Overall architecture of the proposed action recognition model. (a) global grid token, (b) joint heat map, (c) joint map token, (d) STAR-transformer module, and (e) encoder and decoder structure of STAR transformer. An encoder and a decoder have the same structure.}
   \vskip -0.2cm
\label{fig2}
\end{figure*}

\section{Approach}

Figure \ref{fig2} shows the overall structure of the proposed action-recognition model based on the STAR-transformer module. Sixteen video frames and the corresponding skeleton sequences were received as input. Each frame goes through the pre-trained ResNet mixed convolution 18 (MC18) \cite{tran2018closer} to extract local and global feature maps. ResNet MC18 models are unsuitable for the proposed zigzag and binary operations because they reduce the video frame size after an operation. The global feature map, which is the output of the last layer of ResNet MC18, is transformed into a global grid token (GG-token) that represents the visual features of the images (Fig. \ref{fig2} (a)). The local feature map, the output of the middle layer of ResNet MC18, is combined with the joint heat map (Fig. \ref{fig2} (b)) and then transformed into a joint map token (JM-token), as shown in Fig. \ref{fig2} (c). The JM tokens represent the local features of each skeleton joint. The two tokens are aggregated into a multi-class token and then fed into STAR-transformer, as shown in Fig. \ref{fig2} (d), to infer the final action label.

\subsection{Cross-Modal Learning}

We first propose a cross-modal learning method that can combine video frames and skeleton features. The video frames are fed to ResNet MC18, and two feature maps are extracted from the middle and last layers. Because the feature map of the middle layer contains more detailed local features than the last layer, it is used for JM-token extraction, and the last layer is applied for GG-token extraction.

{\bf Global grid token (GG-token):} Let a GG-token $\mathbb{T}_g$ consisting of $\mathit{P}$ tokens be $\mathbb{T}_g^t =\{ \mathit{g}_1^t, ..., \mathit{g}_P^t$\} in video frame $\mathit{t}$. To extract the element of GG-token $\mathit{g}_p^t$ from the $\mathit{t}$-th frame, the input frame is adjusted to a size of 224$\times$224, and the global feature map generated through ResNet MC18 has a size of $\mathit{h}\times\mathit{w}$. The global feature map is again flattened into a vector of size $\mathit{h}\mathit{w}$ ($\mathit{P}$), which becomes the number of elements of $\mathbb{T}_g^t$. Because a global feature map consists of $C$ channels, the number of dimensions of each element of $\mathbb{T}_g^t$  is $\mathit{g}_P^t \in \mathbb{R}^C$. This process continues for every video frame, and thus we can obtain $T$ temporal GG-tokens, as shown in Fig. \ref{fig2} (a).

\begin{figure}[t]
\begin{center}
   \includegraphics[width=1.0\linewidth]{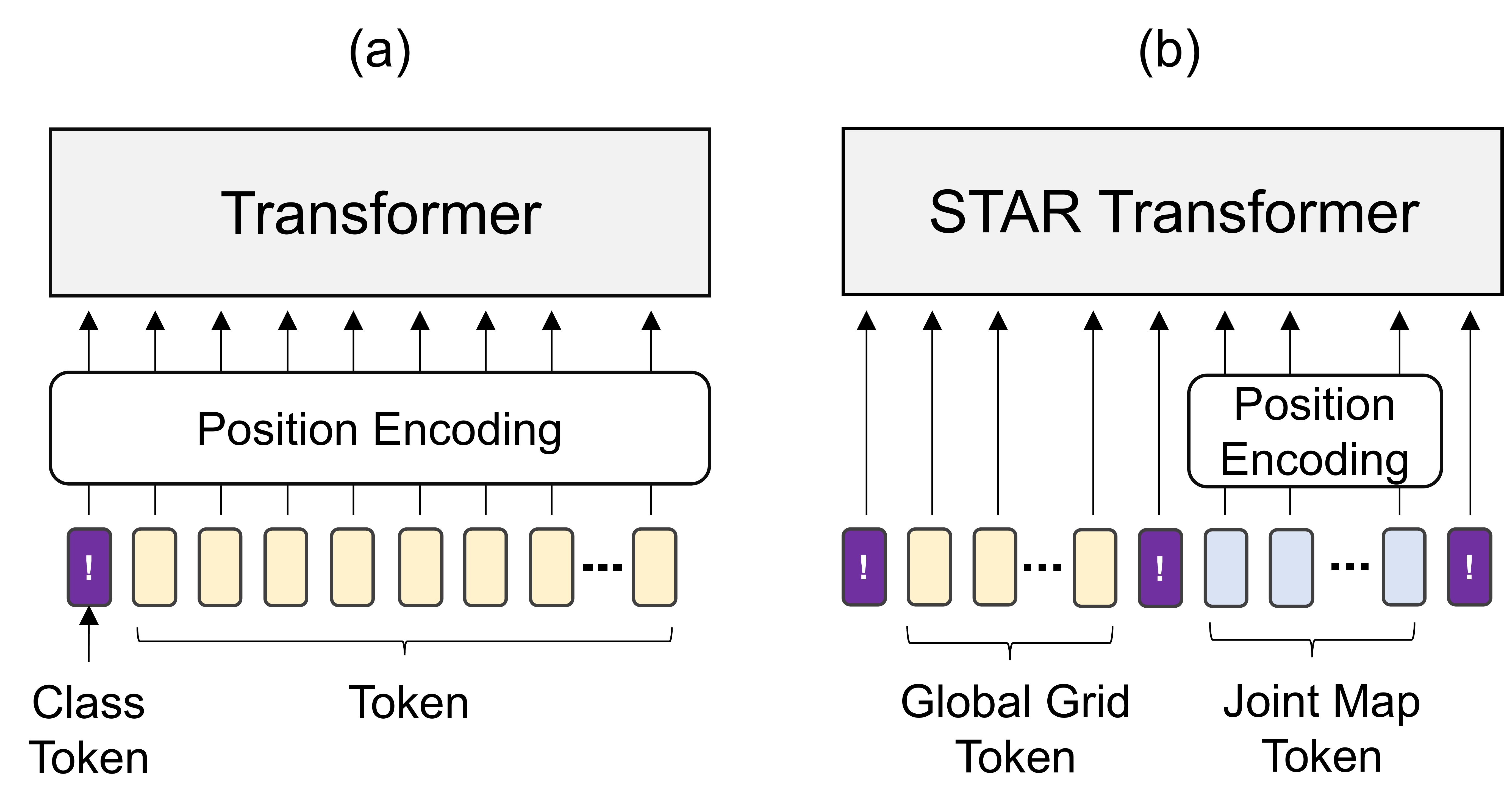}
\end{center}
   \caption{Multi-class token aggregation. (a) single class token generation of pure ViT and (b) proposed multi-token aggregation based on cross-modal learning.}
   \vskip -0.2cm
\label{fig3}
\end{figure}

{\bf Joint map token (JM-token):} In each $t$-th pose corresponding to the $\mathit{t}$-th video frame, we obtain $N$ joint heat maps emphasizing joints for each frame and JM-token set $\mathbb{T}_j^t =\{ \mathit{j}_1^t, ..., \mathit{j}_N^t$\} based on such maps. First, local feature maps $\mathbf{F} \in \mathbb{R}^{\mathit{C}^\prime\times\mathit{h}^\prime\times\mathit{w}^\prime}$ of ResNet MC18 are obtained. The $\mathit{n}$-th joint heat map $\mathit{h}_n \in \mathbb{R}^{\mathit{h}^\prime\times\mathit{w}^\prime}$ is the result of projecting the $\mathit{n}$-th joint onto a temporary map with a size of $\mathit{h}^\prime\times\mathit{w}^\prime$ and applying Gaussian blurring at a scale of $\sigma$. Because a local feature map consists of $\mathit{C}^\prime$ channels, the number of dimensions of each joint element of $\mathbb{T}_j^t$ is $\mathit{j}_N^t \in \mathbb{R}^{\mathit{C}^\prime}$ . The joint element $\mathit{j}_n^t$ on the $\mathit{t}$-th pose is obtained through the concatenation ($\oplus$) of the local feature map F and the $\mathit{n}$-th joint heat map $\mathit{h}_n^t$, as shown in the following equation:

\begin{equation} \label{eqn:my_equation}
\mathbb{T}_{j,n}^{t}=\oplus_{c^\prime=1}^{C^\prime}(\sum_i^{h^\prime}\sum_j^{w^\prime}\mathbf{F}_{c^\prime}(i,j)\times \mathit{h}_n^{t}(i,j) ).
\end{equation}

This process continues for every pose sequence, and thus we can obtain $\mathit{T}$ temporal JM-tokens, as shown in Fig. \ref{fig2} (c).

\begin{figure*}
\begin{center}
    \includegraphics[width=1.0\linewidth]{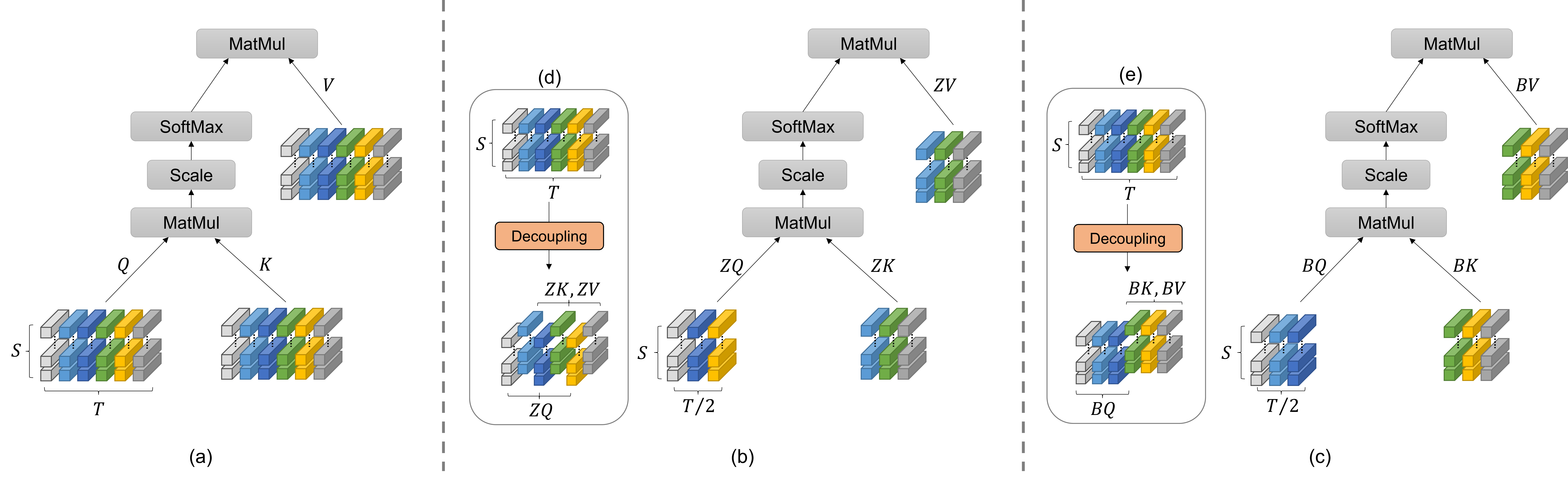}
\end{center}
   \caption{Proposed spatio-temporal cross attention modules. (a) full spatio-temporal attention (FAttn), (b) zigzag spatio-temporal attention (ZAttn), and (c) binary spatio-temporal attention (BAttn) modules.}
   \vskip -0.2cm
\label{fig4}
\end{figure*}

{\bf Multi-class token:} To aggregate the GG- and JM-tokens generated using cross-modal data, we propose a multi-class token aggregation, as shown in Fig. \ref{fig3}. Pure ViT \cite{dosovitskiy2020image}, shown in Fig. \ref{fig3} (a), focuses on learning the global relationships between the input tokens using single-class tokens. However, the proposed action recognition model must cooperatively learn the multi-class tokens generated from the cross-domain data. The proposed aggregation method of multi-class tokens therefore effectively learns the characteristics of different feature representations, as shown in Fig. \ref{fig3} (b).

Multi-class token $\mathbf{Z}$ is created by concatenating ($\oplus$) the class tokens for a GG-token ($\mathrm{CLS}_{glob}$) and JM-token ($\mathrm{CLS}_{joint}$) as follows:

\begin{equation} \label{eqn:my_equation}
\mathbb{T}_{g}=\mathrm{CLS}_{glob}\oplus\mathbb{T}_{g},
\end{equation}
\begin{equation} \label{eqn:my_equation}
\mathbb{T}_{j}=\mathrm{CLS}_{joint}\oplus(\mathbb{T}_{j}+\mathbf{pos}),
\end{equation}
\begin{equation} \label{eqn:my_equation}
\mathbf{Z}=\mathbb{T}_{g}\oplus\mathbb{T}_{j}\oplus\mathrm{CLS}_{total}
\end{equation}

where ${\mathrm{CLS}}_{total}$ is the class token for all tokens. Unlike a GG-token, with a JM-token, the joint position information $\mathbf{pos}$ is important, and thus $\mathbf{pos}$ is only added to the JM-token.

\subsection{Spatio-temporal cross attention}

Inspired by \cite{arnab2021vivit}, we first propose full spatio-temporal attention (FAttn), as shown in Fig. \ref{fig4} (a), which applies the attention mechanism for all tokens within the spatio-temporal dimension. When FAttn is applied to all tokens of time dimension $T$ and spatial dimension $S$, the complexity increases to $O(\mathit{T}^2\mathit{S}^2)$. However, because FAttn alone is insufficient to handle spatio-temporal features, we propose two additional cross-attention mechanisms, \ie, zigzag spatio-temporal attention (ZAttn), as shown in Fig. \ref{fig4} (b), and binary spatio-temporal attention (BAttn), as shown in Fig. \ref{fig4} (c). There is no need to pay attention to all tokens of time dimension $T$. Instead, all tokens are divided into two token groups for ZAttn and BAttn. When ZAttn and BAttn are applied to all tokens of the time dimension $T$ and spatial dimension $S$, the computational complexity is reduced by 0.25-fold in comparison to FAttn with $O(\frac14\mathit{T}^2\mathit{S}^2)$ because the tokens in the time dimension are divided into two groups.

We first obtain the same-sized query ($\mathit{Q}$), key ($\mathit{K}$), and value ($\mathit{V}$) $\in\mathbb{R}^{\mathit{S}\times\mathit{T}}$ matrices from the multi-class token $\mathbf{Z}$ and 
compute the FAttn outputs as follows:

\begin{equation}
\label{equ:my_equation}
\mathrm{\acute{z}}=\mathrm{FAttn}(Q,K,V)
\end{equation}
\begin{equation}
\label{equ:my_equation}
\mathrm{FAttn}(Q,K,V)=\sum_t^T\sum_s^S \mathrm{Softmax}\left\{ \frac{{Q}_{s,t}\cdot {K}_{s,t}}{\sqrt{d_h}} \right\}{V}_{s,t}
\end{equation}

ZAttn learns the detailed process of changing actions. To calculate ZAttn, the odd-numbered vectors in $\mathbf{Z}$ are divided into $ZQ^\prime\in\mathbb{R}^{S\times T/2}$, and the even-numbered vectors in $\mathbf{Z}$ are divided into $ZK^\prime$ and $ZV^\prime\in\mathbb{R}^{S\times T/2}$ in a zigzag manner, as shown in Fig. \ref{fig4} (b). By contrast, the odd-numbered tokens in $\mathbf{Z}$ are divided into $\mathit{ZK}^{\prime\prime}$ and $\mathit{ZV}^{\prime\prime}\in\mathbb{R}^{S\times T/2}$, and the even-numbered vectors in $\mathbf{Z}$ are divided into $\mathit{ZQ}^{\prime\prime}\in\mathbb{R}^{S\times T/2}$.

We calculate $\mathrm{{a}^\prime}\in\mathbb{R}^{S\times T/2}$ and $\mathrm{{a}^{\prime\prime}}\in\mathbb{R}^{S\times T/2}$ individually using the two types of matrices extracted in a zigzag manner using the following formulas, and then concatenate the outputs $\mathrm{{a}^{\prime}}$ and $\mathrm{{a}^{\prime\prime}}$ as the result of ZAttn.

\begin{equation}
\label{equ:my_equation}
\mathrm{{a}^\prime}=\sum_t^{T/2}\sum_s^S \mathrm{Softmax}\left\{ \frac{{ZQ}_{s,t}^\prime\cdot {ZK}_{s,t}^\prime}{\sqrt{d_h}} \right\}{ZV}_{s,t}^\prime\\
\end{equation}
\begin{equation}
\label{equ:my_equation}
\mathrm{{a}^{\prime\prime}}=\sum_t^{T/2}\sum_s^S \mathrm{Softmax}\left\{ \frac{{ZQ}_{s,t}^{\prime\prime}\cdot {ZK}_{s,t}^{\prime\prime}}{\sqrt{d_h}} \right\}{ZV}_{s,t}^{\prime\prime}\\ 
\end{equation}
\begin{equation}
\label{equ:my_equation}
\mathrm{ZAttn}(Q,K,V) = \mathrm{{a}^{\prime}} \oplus \mathrm{{a}^{\prime\prime}}\\
\end{equation}

BAttn is also generated into two groups by dividing the time-dimensional tokens back and forth, as shown in Fig. \ref{fig4} (c). Through this process, it is possible to learn the change at the beginning and end of the action. In the case of BAttn, after dividing $\mathbf{Z}$ into two groups in a binary manner, the front and rear vectors, $\mathit{BQ}^\prime\in\mathbb{R}^{S\times T/2}$ and $\mathit{BK}^\prime$, respectively, and $\mathit{BV}^\prime\in\mathbb{R}^{S\times T/2}$ matrices are calculated. By contrast, the front vectors in $\mathbf{Z}$ are  divided into $\mathit{BK}^{\prime\prime}$ and $\mathit{BV}^{\prime\prime}\in\mathbb{R}^{S\times T/2}$, and the rear vectors in $\mathbf{Z}$ are  divided into $\mathit{BQ}^{\prime\prime}\in\mathbb{R}^{S\times T/2}$. We calculate the individual $\mathrm{{b}^{\prime}}\in\mathbb{R}^{S\times T/2}$ and $\mathrm{{b}^{\prime\prime}}\in\mathbb{R}^{S\times T/2}$ using the two types of matrices with the same formula of ZAttn, and concatenate the output $\mathrm{{b}^\prime}$ and $\mathrm{{b}^{\prime\prime}}$ as the result of BAttn.

\begin{equation}
\label{equ:my_equation}
\mathrm{BAttn}(Q,K,V)=\mathrm{{b}^{\prime}}\oplus \mathrm{{b}^{\prime\prime}}\\
\end{equation}

\subsection{STAR-transformer encoder and decoder}

The proposed STAR-transformer follows a encoder–decoder structure of pure transformer \cite{vaswani2017attention} than pure ViT \cite{dosovitskiy2020image}, as shown in Fig. \ref{fig2} (e). However, the encoder is composed of a series of FAttn (self-attention) and ZAttn $\mathit{L}$ layers, and the decoder is composed of a series of FAttn and BAttn layers. The encoder uses ZAttn to focus on the learning relationships for detailed changes in action, and the decoders use BAttn to learn the relationships for large changes in action.

The structure of the STAR-transformer layer is as follows.

\begin{equation}
\label{equ:my_equation}
\mathrm{\bar{z}_{\mathit{l}}=LN\{ FSTA(\mathrm{{z}_{\mathit{l}-1}})+{\mathrm{{z}_{\mathit{l}-1}}}} \}, l\in\{ 1, 2, ..., L \}
\end{equation}
\begin{equation}
\label{equ:my_equation}
\mathrm{{z}_\mathit{l}^{\prime}}, \mathrm{{z}_{\mathit{l}}^{\prime\prime}}=\mathrm{Decoupling(\bar{z}_{\mathit{l}})}
\end{equation}
\begin{equation}
\label{equ:my_equation}
\mathrm{\tilde{z}_{\mathit{l}}=LN\{ (STA({z}_{\mathit{l}}^{\prime})+{z}_{\mathit{l}}^{\prime})}\oplus \mathrm{(STA({z}_{\mathit{l}}^{\prime\prime})+{z}_{\mathit{l}}^{\prime\prime})} \} 
\end{equation}
\begin{equation}
\label{equ:my_equation}
\mathrm{{z}_{\mathit{l}}=LN\{ MLP(\tilde{z}_{\mathit{l}})+\tilde{z}_{\mathit{l}}} \} 
\end{equation}

Here, $l$ is the number of transformer layers, LN is the layer normalization, and FSTA is the multi-head self-attention for FAttn. Decoupling refers to zigzag or binary grouping. STA represents spatio-temporal attention for ZAttn and BAttn, and MLP is a multi-layer perceptron.

The multi-class tokens output by STAR-transformer are combined into a single class token by averaging and feeding into the MLP to infer the final action label.

\section{Experimental Results}

In this section, we describe the implementation details, including the dataset and training hyperparameters applied. After conducting a quantitative analysis based on SoTA approaches, ablation studies and a qualitative analysis were applied on the effectiveness of multi-expression learning, the number of transformer layers, and spatio-temporal cross attention.

\subsection{Experiment Setup}

{\bf Dataset Description:} The experiment was conducted using the representative action recognition datasets, Penn-Action \cite{zhang2013actemes}, NTU-RGB+D 60 \cite{shahroudy2016ntu}, and 120 \cite{liu2019ntu}. The Penn-Action dataset includes 15 different action classes, such as \textit{baseball swings, jumping jacks,} and \textit{pushups}, for a total of 2,326 RGB video sequences. The NTU-RGB+D 60 dataset is a large dataset used for human action recognition containing 56,880 samples of 60 action classes collected from 40 subjects. Actions are divided into three categories having 40 daily actions (\eg, \textit{drinking, eating,} and \textit{reading}), 9 health-related actions (\eg, \textit{sneezing, staggering,} and \textit{falling}), and 11 mutual actions (\eg, \textit{punching, kicking,} and \textit{hugging}), respectively, based on multi-modal information of the action characterization, including depth maps, 3D skeletal joint positions, RGB frames, and infrared 
sequences. NTU-RGB+D 60 has two evaluation protocols, cross-subject (XSub) and cross-view (XView). NTU-RGB+D 120 extends this version of NTU-RGB+D 60 by adding another 60 classes and containing 114,480 samples in total. NTU-RGB+D 120 has two evaluation protocols, XSub and cross-setup (XSet).



\begin{table}[!t] 
	\begin{center}  
		{\small{
				\begin{tabular}{lcccc}
					\toprule
					\multirow{2}{*}{Method} & \multirow{2}{*}{Pre-train} & \multicolumn{2}{c}{Feature} & \multirow{2}{*}{Acc.} \\
					\cmidrule(lr){3-4}
					&                            & RGB      &   Annot Pose     &                     (\%)  \\
					\midrule
					3D Deep \cite{cao2017body} & \xmark & \cmark & \cmark & 98.1\\
					PoseMap \cite{liu2018recognizing} & \xmark & \cmark & \cmark & 98.2\\
					Multitask CNN \cite{luvizon20182d} & \xmark & \cmark & \cmark &  98.6\\
					HDM-BG \cite{zhao2019bayesian} & \xmark & & \cmark & 93.4\\
					Pr-VIPE \cite{sun2020view} & \cmark & & \cmark &  97.5\\
					UNIK \cite{yang2021unik} & \cmark & & \cmark &  97.9\\
					\midrule
					\rowcolor{Gray}
					STAR-Transformer & \xmark & \cmark & \cmark & 98.7\\
					\bottomrule
				\end{tabular}
		}}
	\end{center}
	\caption{Performance comparison with other state-of-the-art methods on the Penn-Action dataset (Annot, annotated pose(skeleton); Acc, accuracy).}
	\vskip -0.2cm
	\label{tb1}
\end{table}

\begin{table*}[t] 
	\begin{center}
		{\small{
				\begin{tabular}{lcccccccc}
					\toprule
					\multirow{2}{*}{Method} & \multirow{2}{*}{Pre-training} & \multicolumn{3}{c}{Feature} & \multicolumn{2}{c}{NTU60} & \multicolumn{2}{c}{NTU120} \\
					\cmidrule(lr){3-5} \cmidrule(lr){6-7} \cmidrule(lr){8-9}
					& & RGB & Est. Pose & Annot. Pose & XSub & XView & XSub & XSet \\										
					\midrule
					PoseMap \cite{liu2018recognizing} & \xmark & \cmark & \cmark &  & 91.7 & 95.2 & - & - \\
					MMTM \cite{joze2020mmtm} & \xmark & \cmark & & \cmark & 91.9 & - & - & - \\
					VPN \cite{das2020vpn}  & \xmark & \cmark & & \cmark & 95.5 & 98.0 & 86.3 & 87.8 \\
					DualHead-Net \cite{chen2021learning} & \xmark & & & \cmark & 92.0 & 96.6 & 88.2 & 89.3 \\
					Skeletal GNN \cite{zeng2021learning}  & \xmark & & & \cmark & 91.6 & 96.7 & 87.5 & 89.2 \\
					CTR-GCN \cite{chen2021channel} & \xmark & & & \cmark & 92.4 & 96.8 & 88.9 & 90.6 \\
					InfoGCN \cite{chi2022infogcn} & \xmark & & & \cmark & 93.0 & 97.1 & 89.8 & 91.2 \\
					3s-AimCLR \cite{guo2022contrastive} & \cmark & & & \cmark & 86.9 & 92.8 & 80.1 & 80.9 \\
					PoseC3D \cite{duan2022revisiting} & \cmark & \cmark & \cmark & & 97.0 & 99.6 & 95.3 & 96.4 \\
					KA-AGTN \cite{liu2022graph} & \xmark & & & \cmark & 90.4 & 96.1 & 86.1 & 88.0 \\
					\midrule
					\rowcolor{Gray}
					STAR-Transformer & \xmark & \cmark & & \cmark & 92.0 & 96.5 & 90.3 & 92.7 \\
					\bottomrule
				\end{tabular}
		}}
	\end{center}
	\caption{Performance comparison with other state-of-the-art methods on the NTU-RGB+D Action dataset (Est, estimated pose(skeleton); XSub,cross-subject; XView, cross-view; XSet, cross-setup).} 
	\label{tb2}
\end{table*}

{\bf Implementation details:} The proposed STAR-transformer was implemented using PyTorch, and ResNet MC18 pre-trained with Kinetics-400 was applied as the backbone network. When training the model, the Penn-Action and NTU-RGB+D datasets used 16 fixed frames. For all datasets, we utilized a batch size of 4, 300 epochs, an stochastic gradient descent (SGD) optimizer, a learning rate of 2e-4, and a momentum of 0.9. The experiments were conducted in an environment configured with four NVIDIA Tesla V100 GPUs.

\subsection{Comparison with State-of-the-art Methods}

{\bf Penn-Action Dataset:} Table \ref{tb1} shows the results of the comparison experiments with other SoTA action recognition technologies for the Penn-Action dataset: 1) body joint guided 3D deep convolutional descriptors (3D Deep) \cite{cao2017body}, 2) evolution of pose estimation maps (EV-Pose) \cite{liu2018recognizing}, 3) multitask CNN \cite{luvizon20182d}, 4) Bayesian hierarchical dynamic model (HDM-BG) \cite{zhao2019bayesian}, 5) view-invariant 
probabilistic embedding (Pr-VIPE) \cite{sun2020view}, and 6) a unified framework for skeleton-based action recognition (UNIK) \cite{yang2021unik}. UNIK \cite{yang2021unik} was pretrained using the Posetics dataset reconstructed from the Kinect-400 \cite{kay2017kinetics} dataset, and Pr-VIPE \cite{sun2020view} was pretrained using the Human3.6M dataset \cite{ionescu2013human3}. STAR-transformer and the other methods were trained and tested only on the given data, without any pre-training. 

The pre-trained UNIK \cite{yang2021unik} model showed a 0.8\% lower accuracy than the proposed model at 97.9\%, and the Pr-VIPE \cite{sun2020view} model showed 97.5\% accuracy, which is 1.2\% lower than that of the proposed model.

During the experiment, STAR-transformer and the three methods \cite{cao2017body, liu2018recognizing, luvizon20182d} using the RGB of the video frames and the pose (skeleton) feature together showed a high overall performance of 98\% or higher. However, the three methods \cite{zhao2019bayesian, sun2020view, yang2021unik} using only the pose feature showed a relatively low performance of 93\% to 97\%. As the results in Table \ref{tb1} indicate, we can confirm that the action recognition performance can be improved when the RGB of the video frames and pose features are used together. Although STAR-transformer did not use any pre-training, the highest accuracy was derived through the proposed cross-attention using the cross-modal features together.

{\bf NTU-RGB+D Dataset: } Table \ref{tb2} shows the results of the comparison experiments with SoTA action recognition technologies when applying the NTU-RGB+D dataset: 1) long-term localization using 3D LiDARs (PoseMap) \cite{liu2018recognizing}, 2) multimodal transfer module (MMTM) \cite{joze2020mmtm}, 3) video-pose embedding (VPN) \cite{das2020vpn}, 4) multi-granular spatio-temporal graph network (DualHead-Net) \cite{chen2021learning}, 5) skeletal graph neural networks (Skeletal GNN) \cite{zeng2021learning}, 6) channel-wise topology refinement GCN (CTR-GCN) \cite{chen2021channel}, 7) information bottleneck-based GCN (InfoGCN) \cite{chi2022infogcn}, 8) contrastive learning (3s-AimCLR) \cite{guo2022contrastive}, 9) 3D skeleton and heatmap stack (PoseC3D) \cite{duan2022revisiting}, and 10) kernel attention adaptive graph transformer network (KA-AGTN) \cite{liu2022graph}. Because a transformer-based action recognition method that uses RGB of the video frames and cross-modal features of skeleton together has not yet been published, we compare the performance with KA-AGTN, a SoTA for skeleton and transformer-based action recognition.

During this experiment, the accuracy was measured separately for the NTU-RGB+D 60 and NTU-RGB+D 120 
datasets, and the cross-subject (XSub), cross-view (XView), and cross-setup (XSet) were measured separately for each dataset. The performances of four methods \cite{duan2022revisiting, das2020vpn, joze2020mmtm, liu2018recognizing} using the RGB of the video frames and pose together, and six methods using only the pose \cite{liu2022graph, chen2021learning, zeng2021learning, chen2021channel, chi2022infogcn, guo2022contrastive}, were compared with STAR-transformer. Pre-training was conducted using only 3s-AimCLR \cite{guo2022contrastive}. As shown in Table \ref{tb2}, the accuracy was higher for NTU60 and NTU120 when the cross-modal features of the RGB and pose were used together than when a unimodal feature was applied. PoseC3D \cite{duan2022revisiting}  performed 5\% better on NTU60 XSub and 3.7\% better on NTU120 XSet than the proposed STAR-transformer because PoseC3D did not use annotated poses but applied a separate pre-trained poseConv3D model for 3D pose estimation to achieve better action recognition. As the results indicate, PoseC3D achieved a relatively high accuracy because it extracted the optimal pose features suitable for its own model and used them for learning. However, this method still has certain disadvantages in that it requires a pre-trained model for additional pose detection, and the pose detection and action recognition models cannot be trained end-to-end as a single model. KA-AGTN \cite{liu2022graph} used a transformer structure as in our method. However, because it uses only skeleton information and the transformer is used only for spatial information processing between joints, the performance is inferior to the proposed method by up to 1.6\% for NTU60 and up to 4.7\% for NTU120.

Although transformers need to be pre-trained using a large dataset, the proposed STAR-transformer combines the RGB and annotated poses without any pre-training, achieving a promising accuracy even on a larger class dataset NTU120. In particular, NTU120 XSub and XSet showed the second highest performance with accuracy rates of 90.3\% and 92.7\%, respectively. This indicates that STAR-transformer is capable of an excellent action recognition, although the action class is increased or the cross view is changed.

\begin{table}[!t] 
	\begin{center}
		{\small{
				\begin{tabular}{cc}
					\toprule
					Multi-Class token & Accuracy(\%)\\
					\midrule
					\xmark & 97.3\\
					\cmark & 98.7\\
					\bottomrule
				\end{tabular}
		}}
	\end{center}
	\caption{Effectiveness of multi-class token.} 
	\vskip -0.2cm
	\label{tb3}
\end{table}

\subsection{Ablation Study}

In this section, the detailed performance of the modules constituting the proposed STAR transformer model is verified based on several experiments. All experiments were conducted using the Penn-Action dataset.\\

{\bf Effectiveness of multi-expression learning: } To confirm the effect of the proposed multi-expression learning, Table \ref{tb3} presents a comparison experiment conducted with a single-class token used in pure ViT \cite{dosovitskiy2020image} and the multi-class token proposed in this study. The proposed multi-class token performed 1.4\% higher than a single-class token. Although the existing single-class tokens did not effectively conduct learning between the cross-modal tokens, it was confirmed that the proposed multi-class token can effectively increase the performance of the model under the same cross-modal condition.\\

\begin{figure}[t]
\begin{center}
   \includegraphics[width=0.8\linewidth]{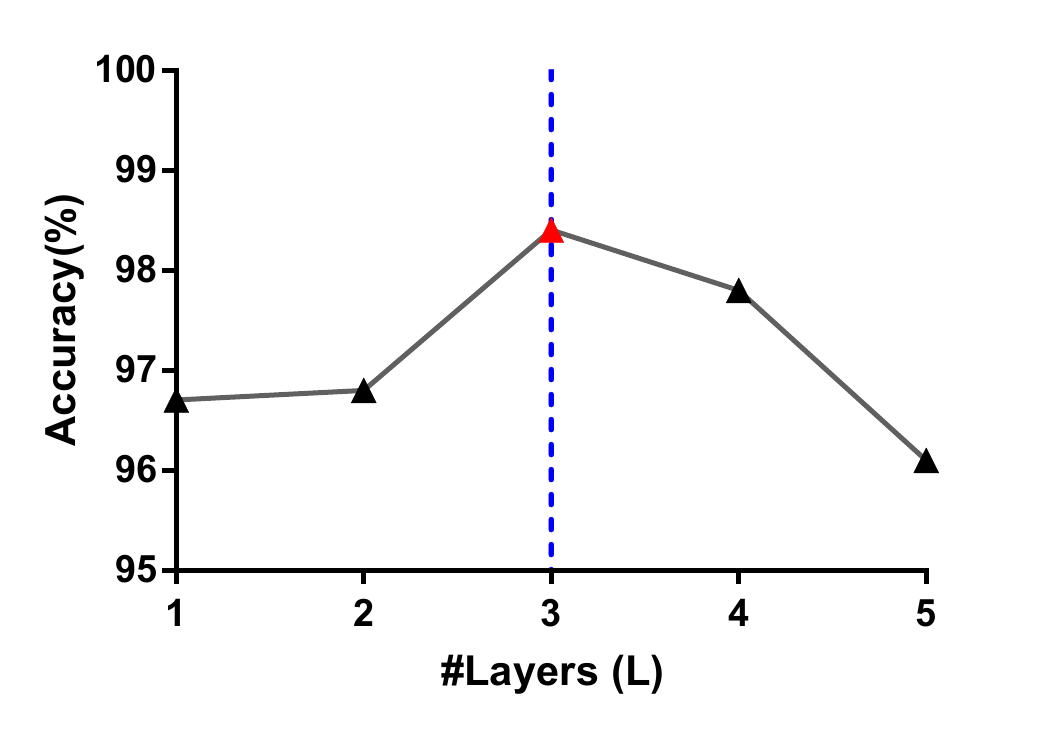}
\end{center}
   \caption{Variation in accuracy according to the number of spatio-temporal cross attention layers.}
\label{fig5}
\end{figure}

{\bf Effectiveness of the number of transformer layers: } Figure \ref{fig5} shows the difference in performance according to the number of transformer layers for the proposed spatio-temporal cross-attention module structure. As shown in Fig. \ref{fig5}, the overall performance improves as the number of layers increases; however, when there are more than four layers, the model is easily overfitted. Therefore, based on the experimental results, we set the number of transformer 
layers to three.\\

\begin{figure*}
\begin{center}
\includegraphics[width=1.0\linewidth]{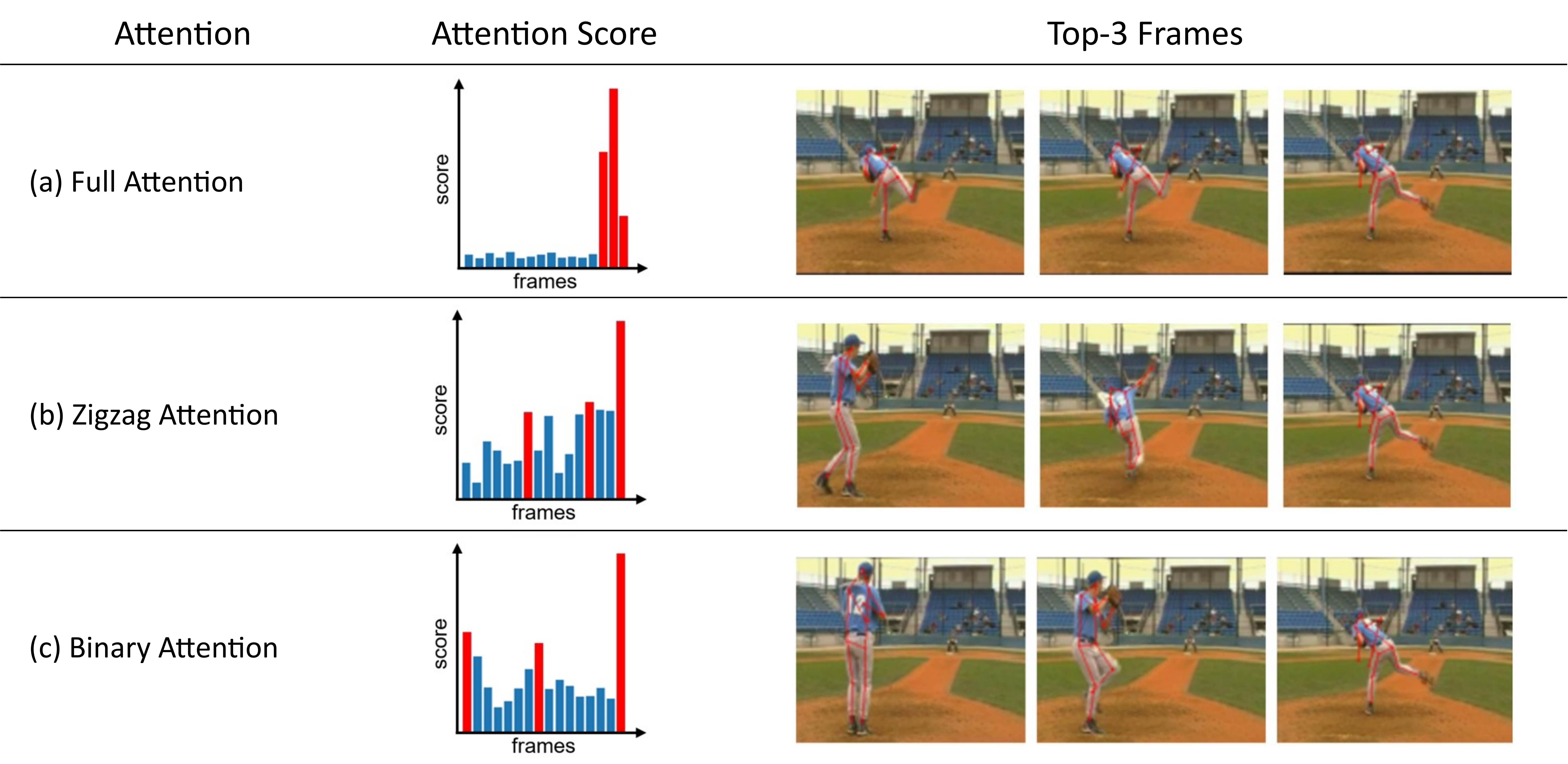}
\end{center}
    
   \caption{The relative importance score of 16 input frames of a validation video. The bar graph shows the attention score for each frame. When only full spatio-temporal attention is used, the attention score appears high at the end of the action. In the case of zigzag spatio-temporal attention, high attention scores were obtained in the middle and last frames when the action was large. In the case of binary spatio-temporal attention, a high attention score appears in the entire frame of an action.}
   \vskip -0.2cm
\label{fig6}
\end{figure*}

{\bf Effectiveness of spatio-temporal cross attention: } Figure \ref{fig6} shows the relative frame importance score for the spatio-temporal cross-attention mechanism proposed in this study. The scores were calculated using an attention rollout \cite{abnar2020quantifying} to calculate the relative concentration for each frame. The attention rollout recursively receives the embedding attention as the input for each layer of the transformer model and computes the token attention.

\begin{table}[!t] 
	\begin{center}
		{\small{
				\begin{tabular}{ccc}
					\toprule
					\multicolumn{2}{c}{Attention Struture} & \multirow{2}{*}{Accuracy (\%)} \\
					\cmidrule(lr){1-2}
					Encoder & Decoder &         \\
					\midrule
					F-F & F-F & 96.1\\
					F-Z & F-Z & 97.3\\
					F-B & F-B & 97.8\\
					F-B & F-Z & 97.3\\
					\rowcolor{Gray}
					\midrule
					F-Z & F-B & 98.7\\
					\bottomrule
				\end{tabular}
		}}
	\end{center}
	\caption{Difference in accuracy between spatio-temporal cross-attention mechanisms. F represents FAttn, Z indicates ZAttn, and B represents the BAttn.}
	\vskip -0.2cm
	\label{tb4}
\end{table}

In Fig. \ref{fig6} (a), FAttn, which has a structure similar to that of ViTs, shows a high score in the last frame, indicating that the front frames are barely considered in action recognition. Here, only the final top-three frames of \textit{after throwing the ball} significantly contributed to the performance, and thus we can see that FAttn does not consider the overall temporal characteristics.

In the case of ZAttn and BAttn, which are spatio-temporal cross-attention mechanisms, the importance scores are equally high in all frames, as shown in Fig. \ref{fig6} (b) and (c). When checking the top-three frames of ZAttn and BAttn, sequentially varied frames such as \textit{before throwing the ball, while throwing the ball,} and \textit{after throwing the ball} are considered for a performance improvement. \\

{\bf Difference in accuracy of cross attention modules: } Table \ref{tb4} shows the differences in accuracy according to the structure of the spatio-temporal cross-attention module. Based on the experimental results, we can see that when ZAttn (Z) and BAttn (B) are used together, the performance is higher than when FAttn is used alone. When FAttn and BAttn were used equally for the encoder and decoder (F-B, F-B), the second highest accuracy was achieved at 97.8\%; however, the importance of the entire frame was still not accurately reflected, and thus the performance was slightly lower than that of the F-Z and F-B combinations. The combination of F-B and F-Z, in which BAttn is applied to 
the encoder and ZAttn is applied to the decoder, showed the second-lowest performance at 97.3\%. Based on the experimental results, we used F-Z as the encoder and F-B as the decoder.

Through these experimental results, we can see that for an accurate action recognition, it is necessary to learn the frame characteristics evenly in all frames through the proposed spatio-temporal cross attention mechanism.

\section{Conclusion}

In this paper, we proposed STAR-transformer, an algorithm based on a spatial-temporal cross-attention module that simultaneously uses video frames and skeleton-based features for action recognition. In addition, the proposed multi-feature representation learning approach was able to flexibly combine the RGB of the video frames, skeleton, and joint trajectories using multi-class tokens. As a result of testing the proposed algorithm using the Penn-Action and NTU-RGB+D action datasets, it was confirmed that the proposed STAR-transformer model achieved substantial improvements in comparison to previous SoTA methods. In a future study, we plan to develop an algorithm that can efficiently learn a model without an overfitting, even with a small number of data. In addition, by extending the proposed STAR-transformer to a model that combines a pose estimation rather than annotated poses, we will modify the STAR-transformer into an end-to-end model that can simultaneously apply pose feature estimation and action recognition optimized for action recognition.





\section*{Acknowledgments}
\thanks{This research was supported by Basic Science Research Program through the National Research Foundation of Korea(NRF) funded by the Ministry of Education(2022R1I1A3058128), and partly supported by the Scholar Research Grant of Keimyung University in 2022.}\\

{\small
\bibliographystyle{ieee_fullname}
\bibliography{egpaper}
}
\end{document}